\documentclass[sigconf,anonymous=false,authorversion=true]{acmart}

\AtBeginDocument{%
  \providecommand\BibTeX{{%
    \normalfont B\kern-0.5em{\scshape i\kern-0.25em b}\kern-0.8em\TeX}}}

\copyrightyear{2022}
\acmYear{2022}
\setcopyright{rightsretained}
\acmConference[]{***}{}{}
\acmBooktitle{***}
\acmDOI{}
\acmISBN{}

\acmSubmissionID{Poster9127}

\usepackage[acronym, style=super, nonumberlist]{glossaries}
\newacronym{DNN}{DNN}{deep neural network}
\newacronym{SGD}{SGD}{stochastic gradient descent}
\newacronym{CG}{CG}{computational graph}
\newacronym{NAS}{NAS}{neural architecture search}
\newacronym{NB101}{NB101}{NAS-Bench-101}
\newacronym{NB201}{NB201}{NAS-Bench-201}
\newacronym{TF}{TF}{training-free}
\newacronym{NASWOT}{NASWOT}{NAS-without-training}
\newacronym{C-BRED}{C-BRED}{Clustering-Based REDuction}
\newacronym{MAC}{MAC}{multiply-accumulate}

\begin{document}

\title[Poster: C-BRED]{Poster: Reducing Neural Architecture Search Spaces with \\ Training-Free Statistics and Computational Graph Clustering}

\author{Thorir Mar Ingolfsson}
\authornote{Both authors contributed equally to this research.}
\affiliation{%
	\institution{ETH Zürich}
	\city{Zürich}
	\country{Switzerland}%
}
\email{thoriri@iis.ee.ethz.ch}

\author{Mark Vero}
\authornotemark[1]
\affiliation{%
	\institution{ETH Zürich}
	\city{Zürich}
	\country{Switzerland}%
}
\email{mveroe@ethz.ch}

\author{Xiaying Wang}
\affiliation{%
	\institution{ETH Zürich}
	\city{Zürich}
	\country{Switzerland}%
}
\email{xiaywang@iis.ee.ethz.ch}

\author{Lorenzo Lamberti}
\affiliation{%
	\institution{Universit\`{a} di Bologna}
	\city{Bologna}
	\country{Italy}%
}
\email{lorenzo.lamberti@unibo.it}

\author{Luca Benini}
\affiliation{%
	\institution{ETH Zürich \& Universit\`{a} di Bologna}
	\city{Zürich, Switzerland \& Bologna}
	\country{Italy}%
}
\email{lbenini@iis.ee.ethz.ch}

\author{Matteo Spallanzani}
\affiliation{%
	\institution{ETH Zürich}
	\city{Zürich}
	\country{Switzerland}%
}
\email{spmatteo@iis.ee.ethz.ch}

\renewcommand{\shortauthors}{Ingolfsson and Vero, et al.}

\begin{abstract}
The computational demands of \gls{NAS} algorithms are usually directly proportional to the size of their target search spaces.
Thus, limiting the search to high-quality subsets can greatly reduce the computational load of \gls{NAS} algorithms.
In this paper, we present \gls{C-BRED}, a new technique to reduce the size of \gls{NAS} search spaces.
\Gls{C-BRED} reduces a \gls{NAS} space by clustering the computational graphs associated with its architectures and selecting the most promising cluster using proxy statistics correlated with network accuracy.
When considering the \gls{NB201} data set and the CIFAR-100 task, \gls{C-BRED} selects a subset with $70\%$ average accuracy instead of the whole space's $64\%$ average accuracy.
\end{abstract}

\ccsdesc[500]{Computing methodologies~Machine learning approaches}

\keywords{deep learning, neural architecture search, training-free statistics, computational graph, clustering}

\maketitle

\section{Introduction}

\Glspl{DNN} have been breaking accuracy records in diverse fields, from computer vision to natural language processing.
However, designing new task-accurate architectures requires considerable expertise, time, and computing resources.
\Gls{NAS} aims at accelerating the development of accurate \glspl{DNN} by parametrising network designs and automating the selection of optimal networks.
The basic working principle of \gls{NAS} is simple: given a collection of candidate network architectures (the \gls{NAS} space), apply a procedure that can select the best performing one (the \gls{NAS} algorithm)~\cite{Elsken2019}.

Most \gls{NAS} approaches have huge computational requirements \cite{Wu2018, Tan2019MnasNet}.
Empirical evidence suggests that both the convergence time and the final performance of \gls{NAS} algorithms depend on the size of the search space and the accuracy distribution of its enclosed networks, with smaller and high-performing spaces having the advantage~\cite{Radosavovic2020, Hu2020, Ci2021}.
Thus, the deep learning community has devoted significant efforts to understanding search spaces and designing increasingly better ones.
However, the proposed techniques either lack full automation or require training large amounts of networks~\cite{Radosavovic2020, Ci2021}.
Some \gls{NAS} algorithms include search space selection in their optimisation loop, but this selection is either not independent of the search algorithm~\cite{Hu2020} or uses extremely simple heuristics~\cite{Lin2020}.

In this work, we provide the following contributions to \gls{NAS}.
\begin{itemize}
    \item We present \gls{C-BRED}, an unsupervised technique to restrict \gls{NAS} spaces to high-performing subsets. \Gls{C-BRED} combines pointwise information (proxy statistics) with relational information (clustering of computational graphs) about network architectures to identify high-quality subsets of the target \gls{NAS} space without requiring any training iteration.
    \item We demonstrate \gls{C-BRED}'s effectiveness on the \gls{NB201} space, where it identifies a subset with $70\%$ average CIFAR-100 accuracy, as opposed to the baseline $64\%$ of the whole space.
\end{itemize}

\section{Background and formulation}

We model the architectures from a given \gls{NAS} space using a latent variable $\lambda \in \Lambda$.
Each architecture manifests itself in two forms: the function form $f_{\lambda}$ and the program form $G_{\lambda}$.

\subsection{Training-free statistics}
In its function form, each architecture is a parametric function $f_{\lambda} \,:\, \Theta_{\lambda} \times X \to Y$, where $X$ is the input space, $Y$ is the output space, and $\Theta_{\lambda}$ is the parameter space.
Given an initial condition $\theta_{\lambda}^{(0)}$, training a network amounts to defining a path $(\theta_{\lambda}^{(0)}, \dots, \theta_{\lambda}^{(T)})$ in the parameter space.
In each state of such a path, statistics about $f_{\lambda}$ can be measured.
The most important statistic is task accuracy, which is measured at the end of training ($\theta_{\lambda} = \theta_{\lambda}^{(T)}$).
Recent \gls{NAS} research has identified \gls{TF} statistics, i.e., statistics that can be measured ahead of training ($\theta_{\lambda} = \theta_{\lambda}^{(0)}$) and correlate with the network's task accuracy.
Examples of \gls{TF} statistics are the condition number of the neural tangent kernel, the number of linear regions cut by a ReLU-activated network in its input space, and the so-called \gls{NASWOT} statistics~\cite{Chen2021, Mellor2021}.

\subsection{Computational graphs}
In its program form, each architecture is a computational graph $G_{\lambda} = (M_{\lambda} \cup K_{\lambda}, R_{\lambda} \cup W_{\lambda})$, where $M_{\lambda}$ is a set of memory nodes, $K_{\lambda}$ is a set of kernel nodes, $R_{\lambda} \subset M_{\lambda} \times K_{\lambda}$ contains read operations, and $W_{\lambda} \subset K_{\lambda} \times M_{\lambda}$ contains write operations.
This interpretation exposes how the information flows through \glspl{DNN} and allows to characterise the workload that they impose on the underlying computing platform~\cite{Liberis2021}.
Note that \gls{DNN} computational graphs can be encoded in different ways, as surveyed in~\cite{White2020}.

\section{Clustering-Based REDuction}

Given a \gls{NAS} space $\Lambda$, our objective is to identify $\Lambda^{*} \subset \Lambda$ such that $|\Lambda^{*}| \ll |\Lambda|$ and the average quality of networks in $\Lambda^{*}$ is better than that of the networks in $\Lambda$.
\Gls{C-BRED} achieves this goal as follows:
\begin{enumerate}
	\item define a \textit{similarity measure} $d(G_{\lambda_{1}}, G_{\lambda_{2}}) \geq 0$ over the program forms $G_{\lambda}$;
	\item cluster $\Lambda$ into a collection $\{ \Lambda_{1}, \dots, \Lambda_{K} \}$ ($K \geq 2$) of candidate subsets using the similarity measure;
	\item measure \gls{TF} statistics for all the program forms $f_{\lambda}$ in each candidate subset;
	\item select $\Lambda^{*} = \Lambda_{k^{*}}$ by means of a \textit{selector function} combining the values of the \gls{TF} statistics of the architectures inside each candidate cluster and comparing the aggregated results.
\end{enumerate}
\Gls{C-BRED} is actually a meta-algorithm in that the graph similarity measure and the cluster selector function can be tuned.

\section{Experimental results}

We evaluated \gls{C-BRED} on the \gls{NB201} benchmark search space.

\Gls{NB201} contains $5380$ unique \textit{cell-based} architectures: each architecture is built by generating a smaller network (the \textit{cell}), replicating it several times, and composing the replicas.
This property allows simplifying computational graph comparisons since we can compare two architectures by comparing their originating cells.
To compare two cells, we defined two similarity measures: the first one is topology-agnostic and compares the frequency with which each kernel type (e.g., convolution, skip connection) is used in the cell; the second one compares instead how many times a given kernel type is used as part of a computational path.
We clustered program forms using the DBSCAN algorithm and empirically found that a combination of the two distances works best.
Our selector function is a heuristic combining all the four \gls{TF} statistics proposed in \cite{Chen2021} and \cite{Mellor2021} (both versions).
We developed such a function by analysing a data set of \gls{TF} statistics that we created for \gls{NB201}.

The average CIFAR-100 accuracy of \gls{NB201} networks is $64\%$, whereas that of the networks in the subset selected by \gls{C-BRED} is $70\%$.
To validate this preliminary evaluation, we compared \gls{C-BRED} to two subspace selection alternatives.
The alternatives we chose partition \gls{NB201} into five subsets according to the $20\%$ quantiles associated with two statistics: the \gls{NASWOT}v2 statistic and the number of \gls{MAC} operations.
We then used the \gls{NASWOT} technique introduced in \cite{Mellor2021} as the reference search algorithm.
As can be seen in Table~\ref{tab:nb201}, \gls{C-BRED} has superior performance in that not only the average network accuracy is better, but it is also more stable ($\sim 4 \times$ decrease in standard deviation).

\begin{table}
	\caption{CIFAR-100 accuracy distribution of the networks selected by the $10$-run NASWOT algorithm~\cite{Mellor2021} when applied to different subsets of the \gls{NB201} benchmark search space.}
    \centering
	\begin{tabular}{@{}lrrrr@{}}
		\toprule
							& Intrinsic & TF-Q  & MAC-Q & C-BRED 		 \\
		\midrule
		Mean 				& 69.18 	& 69.54 & 69.86 & \textbf{70.34} \\
		Median 				& 69.60 	& 69.90 & 70.16 & \textbf{70.23} \\
		Standard deviation  &  1.93 	&  1.74 &  1.47 & \textbf{0.41}  \\
		\bottomrule
	\end{tabular}
	\label{tab:nb201}
\end{table}

\section{Conclusions}

In this paper, we presented \gls{C-BRED}, a technique to identify subsets of \gls{NAS} spaces containing high-quality architectures.
\Gls{C-BRED} uses relational information about computational graphs to cluster the target \gls{NAS} space into a collection of candidate subsets; then, it uses pointwise information (\gls{TF} statistics) to identify the most promising cluster without requiring any training iteration.


\end{document}